%% file: main.tex
\pdfoutput=1
\documentclass[10pt,twocolumn,letterpaper]{article}

\usepackage{cvpr}              
\usepackage[accsupp]{axessibility}
\usepackage{multirow}
\usepackage{times}
\usepackage{cite}
\usepackage{amsmath,amssymb,amsfonts}
\usepackage{algorithmic}
\usepackage{graphicx}
\usepackage{textcomp}
\usepackage{color}
\usepackage{comment}
\usepackage{enumitem}

\usepackage[pagebackref,breaklinks,colorlinks]{hyperref}

\usepackage[capitalize]{cleveref}
\crefname{section}{Sec.}{Secs.}
\Crefname{section}{Section}{Sections}
\Crefname{table}{Table}{Tables}
\crefname{table}{Tab.}{Tabs.}

\def\cvprPaperID{52} 
\def\confName{CVPR}
\def\confYear{2023}

\begin{document}

\title{A Data-Centric Solution to NonHomogeneous Dehazing via Vision Transformer}



\author{Yangyi Liu$^1$, Huan Liu$^1$, Liangyan Li$^1$, Zijun Wu$^2$ and Jun Chen$^1$\\
$^1$McMaster University, Hamilton, Canada\\%
    $^2$China Telecom Research Institute, Shanghai, China\\%
    \textit {\{liu5, lil61, chenjun\}@mcmaster.ca}, \textit{liuh127@outlook.com}, \textit{wuzj12@chinatelecom.cn}
}

\twocolumn[{
\maketitle
\vspace{-8mm}
\begin{center}
    \captionsetup{type=figure}
    \includegraphics[width=1.0\textwidth]{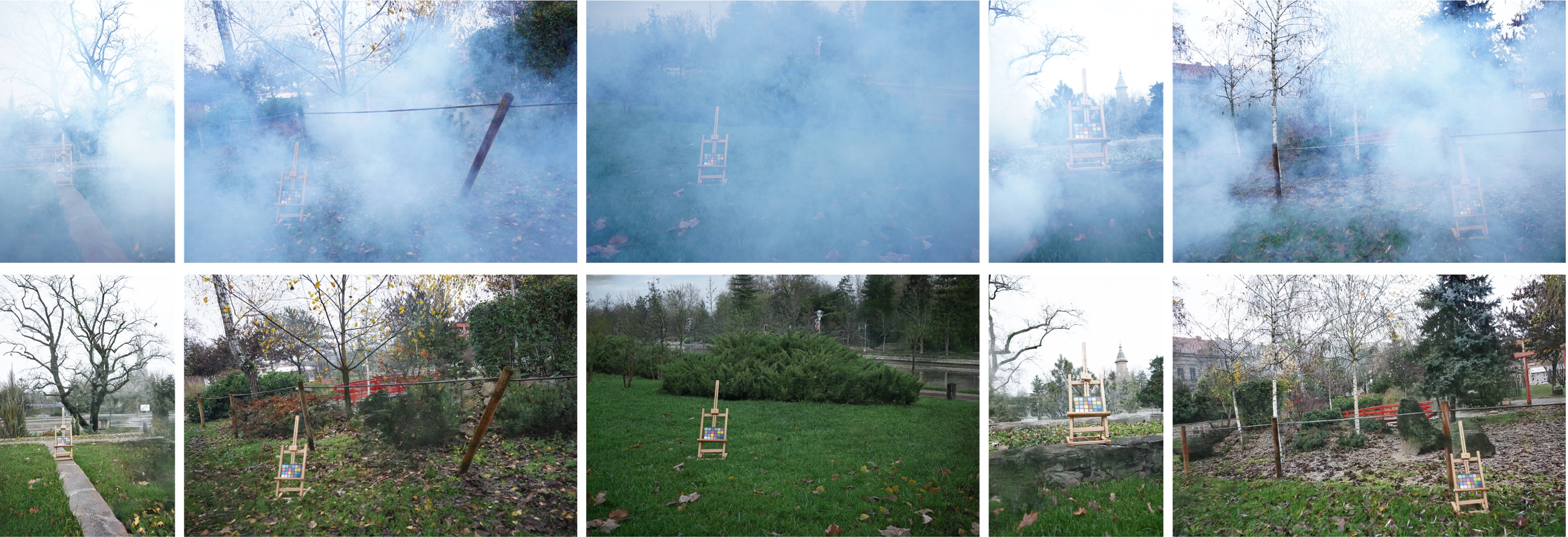}
    \captionof{figure}{Our results on NTIRE 2023 dehazing challenge, achieving the best performance in terms of PNSR, SSIM and LPIPS.}
    \label{Fig.first_page}
\end{center}
}]
\input{abstract}
\input{introduction}
\input{related_works}

\input{methods}

\input{experiments}
\input{conclusion}


{\small
\bibliographystyle{ieee_fullname}
\bibliography{egbib}
}

\end{document}

%% file: abstract.tex
\begin{abstract}
Recent years have witnessed an increased interest in image dehazing.
Many deep learning methods have been proposed to tackle this challenge, and have made significant accomplishments dealing with homogeneous haze.
However, these solutions cannot maintain comparable performance when they are applied to images with non-homogeneous haze, e.g., NH-HAZE23 dataset introduced by NTIRE challenge.
One of the reasons for such failures is that non-homogeneous haze does not obey one of the assumptions that is required for modeling homogeneous haze.
In addition, a large number of pairs of non-homogeneous hazy image and the clean counterpart is required using traditional end-to-end training approaches, while NH-HAZE23 dataset is of limited quantities.
Although it is possible to augment the NH-HAZE23 dataset by leveraging other non-homogeneous dehazing datasets, we observe that it is necessary to design a proper data-preprocessing technique that reduces the distribution gaps between the target dataset and the augmented one.
This finding indeed aligns with the essence of data-centric AI.
With a novel network architecture and a principled data-preprocessing approach that systematically enhances data quality, we present an innovative dehazing method.
Specifically, we apply RGB-channel-wise transformations on the augmented datasets, and incorporate the state-of-the-art transformers as the backbone in the two-branch framework.
We conduct extensive experiments and ablation studies to demonstrate the effectiveness of our proposed method.
The source code is available at \url{https://github.com/yangyiliu21/ntire2023_ITBdehaze}.
\end{abstract}

%% file: introduction.tex
\section{Introduction}
Recent years have witnessed an increased interest in image dehazing, which is categorized as one of the sub-tasks in image restoration.
Haze naturally exists all over the world, and has become more frequent due to the climate change.
This common atmospheric phenomenon has drawn significant attention because of its potential risks to traffic safety, as both the human observation and computer vision models are prone to fail in hazy scenes.
These make image dehazing an important low-level vision task, and many methods have been proposed to tackle this challenge \cite{dcp,li2017aod, cai2016dehazenet, zhang2018densely, Wu_2020_CVPR_Workshops, wu2020knowledge,Qin_Wang_Bai_Xie_Jia_2020,  Liu_2020_CVPR_Workshops, zhou2020cggan, ren2018gated, yu2021two, Fu_2021_CVPR, liu2022towards}.


Among them, many neural network based approaches \cite{cai2016dehazenet, li2017aod, GCA, Qin_Wang_Bai_Xie_Jia_2020, zhang2018densely, zhou2020cggan} show remarkable performance in handling image dehazing problem.
Specifically, benefiting from powerful network modules and vast training data, the end-to-end approaches deliver promising results.
However, as the distribution of haze becomes more complicated and non-homogeneous, many of them fail to achieve satisfying results.
The reason for such failures is because the thickness of the  non-homogeneous haze is not determined entirely by the depth of the background scene. 


Although researchers have made tremendous efforts collecting data with non-homogeneous haze, e.g., the NH-HAZE datasets \cite{ancuti2020ntire, ancuti2021ntire, ancuti2023ntire}, the quantity is still limited.
A common belief is that models are prone to encounter the overfitting problems when training a deep neural network from scratch with such small datasets.
A naive solution is to combine all the available non-homogeneous haze datasets together to form a relatively larger dataset.
However, due to the differences between datasets caused by a variety of factors, such as color distortion, objects complexity and camera capability, it has been shown that a direct combination actually compromises the dehazing performance on individual datasets \cite{liu2022towards}.
It remains a serious challenge to find a robust solution to cope with the practical situation where both the quality and quantity of the available data limited.

To address the above-mentioned problems, we adopt the two-branch framework consisting of state-of-the-art backbone networks, with a novel data-preprocessing transformation applied on the NH-HAZE datasets from previous years.
Motivated by the idea of data-centric AI that machine learning has matured to a point that high-performance model architectures are widely available, while approaches to engineering datasets have lagged \cite{motamedi2021datacentric, dc_comp}, we put much effort in engineering the data.
Inspired by the promising performance of gamma correction \cite{yu2021two,Fu_2021_CVPR}, we propose a simple yet effective RGB-channel-wise data-preprocessing approach.
We demonstrate its suitability for this competition setting, and argue that it is prospective to be the principle for augmenting similar dataset.
Details of this data-centric AI inspired preprocessing approach are discussed in later sections.
Regarding to the network architecture, we design our model under the two-branch framework \cite{yu2021two,Fu_2021_CVPR,wu2020knowledge}.
In the first branch, we adopt the Swin Transformer V2 model \cite{liu2022swin} pre-trained on ImageNet dataset \cite{deng2009imagenet} as the encoder.
The powerful Swin Transformer is accredited to be able to supersede the previous methods in many contexts of transfer learning, where the knowledge gained from large-scale benchmark is adapted to task-specific datasets \cite{Kornblith_2019_CVPR, liu2022swin}. Such pertinent features are of vital importance when dealing with small real-world non-homogeneous datasets \cite{yu2021two}. 
Alongside a refined decoder and skip connections, the first branch extracts multi-level features of the hazy images. 
The second branch is introduced to complement the knowledge learned from the pre-trained model by exclusively working on the domain of target data. 
For simplicity, we follows \cite{yu2021two} to build the second branch with a RCAN \cite{zhang2018image}.
Since there is no down-sampling and up-sampling operations in the second branch, we expect it to extract features distinct from the ones obtained by the first branch.
Finally, a fusion tail aggregates the results from both branches and produces dehazed output images.

Overall, our contributions are summarized as follows.
Firstly we put forward a simple but effective data-preprocessing approach inspired by data-centric AI, leveraging extra data to significantly enhance our model. 
Secondly, we incorporate the state-of-the-art backbone in the two-branch framework. 
By carefully balancing the two branches, our model demonstrates promising results using limit-sized datasets, and outperforms other current approaches adopting this pipeline.
Finally, we conduct extensive experiments to demonstrate the competitive performance of our proposed method. 
With substantial ablation study on different combinations of models and data, we hope to convince the future competition participants to pay equal attention to model design and data engineering.

%% file: related_works.tex
\begin{figure*}[!t]
\centering
\includegraphics[width=0.9\textwidth]{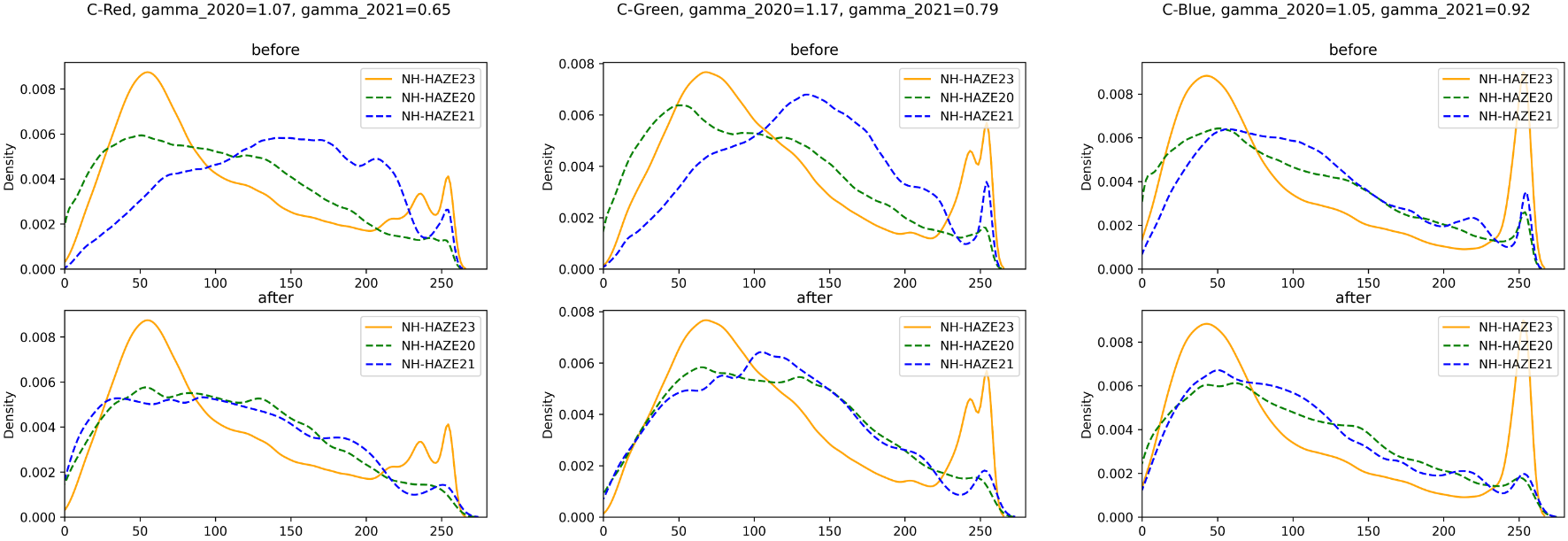}
\caption{
Comparison of RGB-wise distribution of datasets (GT) before and after being processed by our proposed method.}
\label{Fig.clean_hist}
\end{figure*}

\section{Related Works}

In this section, we briefly review the literature of single image dehazing and learning with limited data.

\textbf{Single Image Dehazing}.
Approaches proposed for single image dehazing are divided into two categories: prior-based methods and learning-based methods.
To guarantee the performance, prior-based methods require reasonable assumptions and knowledge on hazy images to obtain accurate estimations about the transmission map and atmospheric light intensity in ASM modeling \cite{middleton1952vision}.
Representative works in this category include \cite{tan2008visibility, dcp, zhu2015fast, fattal2014dehazing, berman2016non}.
Specifically, \cite{tan2008visibility} observed that clear images have higher contrast comparing to the hazy counterparts, and proposed a local contrast maximization method.
Based on the assumption that image pixels in no-haze patches have intensity values close to zero in at least one color channel, \cite{dcp} introduced Dark Channel Prior (DCP).
\cite{zhu2015fast} presented a linear model adapting color attenuation prior (CAP) to estimate the depth according to the knowledge about the difference between the brightness and the saturation of hazy images.
Prior-based methods left a permanent mark in single image dehazing but their vulnerability when adapted in variable scenes pivoted the researchers to another direction, the learning-based methods.
With the advances in neural networks, \cite{cai2016dehazenet, li2017aod, GCA, Qin_Wang_Bai_Xie_Jia_2020, zhang2018densely, zhou2020cggan} have proposed progressively more powerful models that are capable of directly recovering the clean image from hazy image without estimating the transmission map and depth.
The superiority of these methods in removing homogeneous haze is attributed to the availability of large training datasets. 
When applied on non-homogeneous haze, they fail to yield comparable results. 
The limited quantity of existing non-homogeneous haze datasets prevents researchers from adopting simple end-to-end training methods.


\textbf{Learning with Limited Data}.
Data is indispensable for all the AI models.
Many of the models demand a huge dataset for training, but large dataset is not always available.
Therefore, it urges the researchers to find solutions to accomplish training with limited data.
In terms of dehazing, a seemingly straightforward solution to address the issues caused by small non-homogeneous training datasets is composing a relatively large dataset by combining several small datasets all together.
In terms of NTIRE2023 challenge \cite{ancuti2023ntire}, it can be done by augmenting the NH-HAZE datasets (augmented dataset) \cite{ancuti2020ntire, ancuti2021ntire} with this year's data (target dataset). 
Surprisingly, against the common believe that larger dataset is always better in deep learning, \cite{liu2022towards} observed that the models perform better when training and testing are conducted on a single dataset (as opposed to the union of all datasets).
This observation indicates that the augmented dataset locates in a different domain comparing to the target data.
Direct aggregation introduces domain shift problem within the dataset.
Thereby, \cite{liu2022towards} proposed a testing time training strategy to mitigate the problems, while \cite{yu2021two, Fu_2021_CVPR, shao2020domain} chose to adjust the domains of training data before sending them into the dehazing modules.
Interestingly, the idea of focusing on improving the dataset rather than the model was introduced by the Data-Centric AI competition \cite{dc_comp}.
Data-Centric AI is anticipated to deliver a set of approaches for dataset optimization, thereby enabling deep neural networks to be effectively trained using smaller datasets \cite{motamedi2021datacentric}.
The set of proposed techniques ranges widely from simple ones to complex combinations \cite{zha2023datacentric}. 
Through our experiments and qualitative analysis, we find that a too simple approach, such as the gamma correction adopted by \cite{yu2021two, Fu_2021_CVPR} fails to recover the color accurately.
Nevertheless, a complicated method, like \cite{shao2020domain} applying domain adaptation to learn a separate neural network to translate the data, is infeasible due to the scarcity and lacking of depth information of the available data.
In the next section, we introduce our innovative solution standing out in the NTIRE challenge settings.

%% file: methods.tex
\begin{figure*}[!t]
\centering
\includegraphics[width=0.98\textwidth]{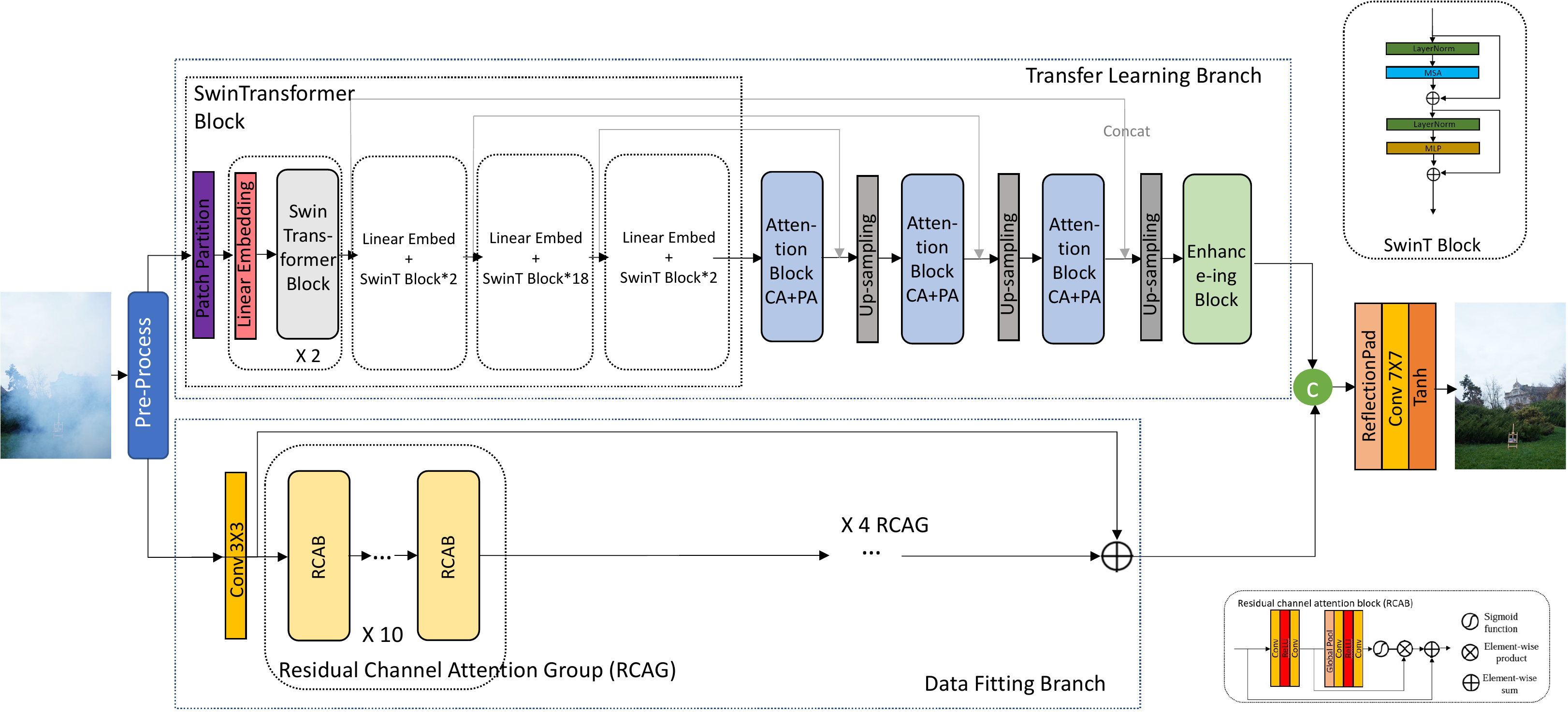}
\caption{An overview of our network. The model consists of two branches.  The transfer learning branch is composed by Swin Transformer based model. The data fitting branch consists of residual channel attention groups.}
\label{Fig.main}
\end{figure*}

\section{Proposed Method}
In this section, we introduce the details of our methodology following the order of the working pipeline.
Firstly, we demonstrate the data-preprocessing method inspired by the idea of data-centric AI.
Secondly, details of our model architecture are presented, as well as the functions of each component.
Finally, we introduce the loss functions applied to train our proposed networks.

\subsection{Data-Centric Engineering}
\label{sec:dpp}
Systematically engineering the data is a key requirement for training deep neural networks.
The idea of data-centric AI moreover emphasizes on assessing the data quality before deployment \cite{zha2023datacentric}.
By comparing the NH-HAZE20 and 21 dataset \cite{ancuti2020ntire, ancuti2021ntire} to the data provided this year both numerically and empirically, we notice obvious color discrepancy.
When evaluating on this year's test data, training on a direct combination of all data does not boost the score comparing to training on this year's data only (see results in Section \ref{sec:data_centric}).
Therefore, our goal is to propose an approach that reduces the color differences, and shifts the distribution of augmented data towards that of target data.
%
Inspired by the success application of gamma correction \cite{yu2021two,Fu_2021_CVPR} as a simple yet effective data-preprocessing technique, 
we propose a more systematic solution for data engineering.
Instead of the practice in \cite{yu2021two,Fu_2021_CVPR} by applying gray-scale gamma correction, we here introduce to correct on each R,G,B channel separately:
\begin{equation} \label{GC}
O_{R,G,B} = (\frac{I_{R,G,B}}{255}) ^{\frac{1}{\gamma_{R,G,B}}}
\end{equation}
where $O$ and $I$ are output and input pixel intensity ($\in [0, 255]$), respectively. $\gamma$ is the gamma factor. The subscripts $R,G,B$ indicate that the values for different channels are unique.

As for implementation, we first calculate the average pixel intensity of each channel of the three datasets; then for each channel of the NH-HAZE20 or 21 dataset, we apply a transformation with a unique gamma value to all the pixels, resulting in similar mean and variance values comparing with the corresponding channel of NH-HAZE23 dataset.
In Figure \ref{Fig.clean_hist}, we present the histogram change with corresponding $\gamma$ values.\
From observation, our method adjusts the color of NH-HAZE20 and 21 data to become much similar to the NH-HAZE23 data.
Numerically, the average pixel intensity of 2023 data is 107.46(R), 114.48(G), 101.92(B). 
After applying our method, the adjusted average pixel intensity of NH-HAZE20 data is 107.77(R), 114.33(G), 102.08(B); and the adjusted ones of NH-HAZE21 data is 107.43(R), 115.01(G), 102.13(B).
Note that, we not only apply such preprocessing method on the clean ground truth images but also on the hazy images (as opposed to \cite{yu2021two,Fu_2021_CVPR} only manipulating the ground truth images).

With this novel data-preprocessing method, the distributions of all three color channels of NH-HAZE20 and 21 data are shifted closer to those of NH-HAZE23 dataset.
Benefiting from more in-distribution data, the models gain substantial improvements.
Being able to work with small but good dataset, rather than a larger but internally diverged one helps us stand out in the competition.
This indeed aligns with the idea of data-centric AI \cite{zha2023datacentric, motamedi2021datacentric}.
For future competition participants, we elect this approach to be a good starting point for data engineering.

\subsection{Network Architecture}
\label{sec:model_arch}
As shown in Figure \ref{Fig.main}, the pre-processed data is fed into a two-branch model architecture. 
This two-branch framework has been successfully employed in various computer vision tasks \cite{jacobsadaptive}, and has facilitated several works \cite{yu2021two,Fu_2021_CVPR,wu2020knowledge} winning the awards in the past NTIRE challenges.
In our implementation, the first Transfer Learning Branch aims to extract pertinent features of the inputs with pre-trained weights initialization.
The second Data Fitting Branch is responsible to complement the knowledge learned from the first branch and work on the domain of target data. 
The fusion tail aggregates the outputs from both branches and produces dehazed images.

\textbf{Swin Transformer based Transfer Learning}.
To leverage the power of transfer learning \cite{tan2018survey}, we use the ImageNet \cite{deng2009imagenet} pre-trained Swin Transformer \cite{liu2022swin} as the backbone of our encoder.
Swin Transformer achieves the state-of-the-art performance in many vision tasks. 
It is exceptionally efficient and more accurate as comparing to its predecessor, Vision Transformer (ViT) \cite{dosovitskiy2020image}, which struggles with high resolution images because its complexity is quadratic to the input size.
The working pipeline of the Swin Transformer is summarized as follows.
First, Swin Transformer splits an input image into non-overlapping patches with a patch splitting module.
Through a linear embedding layer, the patches and their features are set as a concatenation of the raw pixel RGB values, also referred to as “token”, and then be projected to an arbitrary dimension.
These tokens are processed by a cascade of stages.
Each stage consists of a linear embedding layer and several Swin Transformer Block (SwinT Block) modules.
SwinT Block uses cyclic-shift with MSA modules to implement efficient batch computation for shifted window partitioning.
From the previous stage to the next, the spatial dimension of the feature maps are effectively reduced, resulting in hierarchical feature maps.
These modules compose our encoder part of the Transfer Learning Branch.
As for the decoder part, we adopt the ideas from \cite{yu2021two, Fu_2021_CVPR}. 
With skip connections, the attention blocks and up-sampling layers gradually restore the hierarchical feature maps and produce an output with the same spatial dimension as the input.

\textbf{Rest of the Model}.
We adopt the Data Fitting Branch from \cite{zhang2018image}, which is based on residual channel attention block \cite{zhang2018image}.
Trained from scratch, this second branch complements the first one by exclusively working on the domain of target data.
With no down-sampling and up-sampling operations, this branch operates in the full-resolution mode,  thus extracts features distinct from the ones obtained by the first branch.
A simple yet insightful fusion tail consisting of a reflection padding layer, a 7 $\times$ 7 convolutional layer and the Tanh activation \cite{yu2021two} combines the features from two branches and produces dehazed images.

\subsection{Loss Functions}
Since our method mainly focuses on the data-centric engineering and implementing transformers, we follow \cite{yu2021two,Fu_2021_CVPR} to adopt a combination of several losses for training our model. 

\textbf{Smooth L1 Loss}.
For image fidelity reconstruction, the smooth L1 loss \cite{Girshick_2015_ICCV} has been proved to be more robust than the MSE loss in various image restoration tasks \cite{zhao2016loss}. The formulation follows:
\begin{equation}
    L_{l_1} = \frac{1}{N}\sum_i^Nsmooth_{L_1}(y_i - f_\theta(x_i)),
\end{equation}
\begin{equation}
    smooth_{L_1}(z) = \begin{cases}
0.5z^2 & \text{if }  |z| < 1\\
|z| - 0.5& \text{otherwise}
,
\end{cases}
\end{equation}
where $x_i$ and $y_i$ denote the  $i$-th pixel of clean and hazy images, respectively. $N$ is the total number of pixels. $f_\theta(\cdot)$ represents the network.

\textbf{MS-SSIM Loss}.
Multi-scale Structure similarity (MS-SSIM) is based on the assumption that human eyes are adapted for extracting structural information, and therefore a metric of evaluating structural similarity can provide a good approximation to perceived image quality.
Let $O$ and $G$ represent two windows centered at the $i$-th pixel in the dehazed image and the ground truth image, respectively.
Gaussian filters are applied on both windows, and produce resulting values of corresponding means ($\mu_O$, $\mu_G$), standard deviations $\sigma_O$, $\sigma_G$, and covariance $\sigma_{OG}$.
The SSIM formulation for the $i$-th pixel follows:
\begin{equation}
\begin{aligned}
    \text{SSIM}(i)&=\frac{2\mu_O\mu_G + C_1}{\mu_O^2 + \mu_G^2 +C_1} \cdot \frac{2\sigma_{OG} + C_2}{\sigma_O^2 + \sigma_G^2 +C_2}  ,
    \label{con:1}
\end{aligned}
\end{equation}
where $C_1$ and $C_2$ help stabilize the division.

\textbf{Perceptual Loss}.
Besides pixel-scale supervision on perceptual quality, we adopt ImageNet \cite{deng2009imagenet} pre-trained VGG16 \cite{simonyan2014very} to measure perceptual similarity, which helps reconstruct finer details \cite{zhu2017unpaired}.
Denoting $x$ and $y$ as hazy inputs and ground truth images respectively, the loss is defined as:
\begin{equation}
    L_{perc} = \frac{1}{N}\sum_{j}\frac{1}{C_jH_jW_j}||\phi_j(f_\theta(x)) - \phi_j(y)||_2^2 ,
\end{equation}
where $f_\theta(x)$ is the dehazed image. $\phi_j(\cdot)$ denotes the feature map. We choose $L_2$ loss to measure the distances between them. $N$ denotes the number of features. 

\textbf{Adversarial Loss}.
Compensating for the risks that pixel-wised loss functions fail to provide sufficient supervision when training on a small dataset, we employ the adversarial loss \cite{zhu2017unpaired}:
\begin{equation}
    L_{adv} = \sum_{n = 1}^N - logD(f_\theta(x)),
\end{equation}
where $f_\theta(x)$ denotes the dehazed image. $D(\cdot)$ represents the discriminator.

\textbf{Total Loss}.
The total loss is a weighted sum of afore-mentioned four components with pre-defined weights:
\begin{equation}
   L = L_{l_1} + 0.5 L_{\textit{MS-SSIM}} + 0.01 L_{perc} + 0.0005 L_{adv}.
\end{equation}

%% file: experiments.tex
\begin{figure}[!t]
\centering
\includegraphics[width=0.9\linewidth]{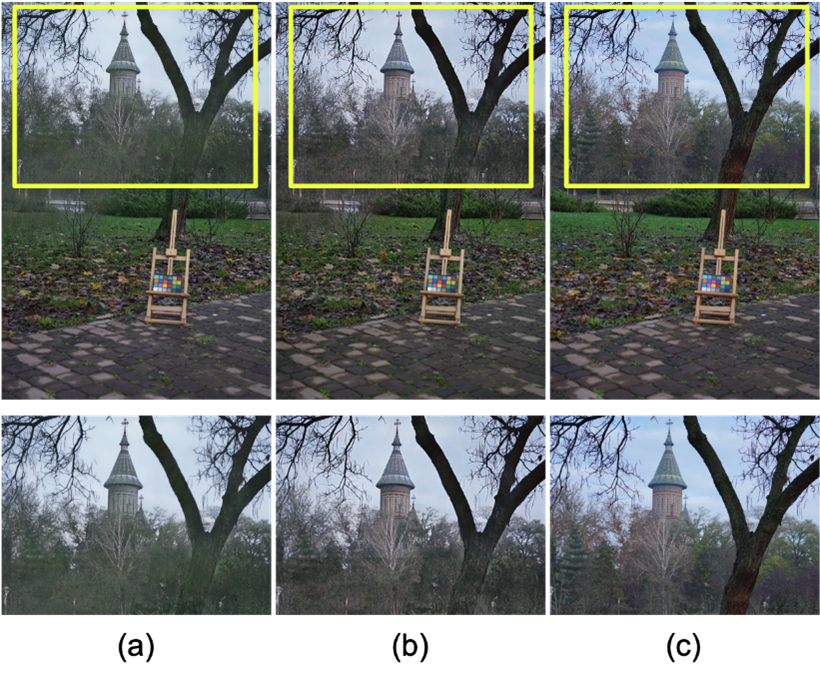}
\caption{ Qualitative ablation study on the data-centric design. (a) the results generated by the model trained on NH-HAZE20+21+23 GC, (b) the results generated by the model trained on NH-HAZE20+21+23 RGB, and (c) ground truth.
}
\label{Fig.ablation}
\end{figure}

\section{Experiments}
In this section, we first introduce the datasets used to conduct experiments along with implementation details.
Then, we conduct ablation studies to verify the effectiveness of our model design and data-preprocessing method.
Finally, we evaluate the dehazing results of our proposed method qualitatively and quantitatively, and compare with several state-of-the-art methods.

\subsection{Datasets}

\textbf{O-HAZE}.
With the help of the professional haze machine that generate real haze, O-HAZE \cite{O-HAZE_2018} was published in 2018, containing 45 clean and hazy image pairs in total.
Each pair has a unique spatial resolution.
We conduct our evaluation based on the official train, and test split \cite{ancuti2018ntire}.

\textbf{DENSE-HAZE}.
DENSE-HAZE \cite{Dense-Haze_2019, dense-haze_report} was introduced with the NTIRE2019 challenge.
It characterizes in dense and homogeneous haze.
The dataset contains 45 training images, 5 validation images and 5 test images.
All images are of the same 1600$\times$1200 dimension.
In our experiments, we follow the official train, val and test split.

\textbf{NH-HAZE20 \& NH-HAZE21}.
In NTIRE2020 \cite{ancuti2020ntire} and NTIRE2021 \cite{ancuti2021ntire} challenges, NH-HAZE20 and NH-HAZE21 were released. 
The haze pattern in these two datasets is non-homogeneous.
The images in these two datasets are of the size of 1600 $\times$ 1200.
NH-HAZE20 contains 45 training data, 5 validation data and 5 testing data.
We adopt the official train, and test split to conduct experiments on NH-HAZE20.
For NH-HAZE21, we take the first 20 training images as our training set, and  the rest 5 images are used for testing.

\textbf{NH-HAZE23}.
\label{subsec:data}
Inheriting the non-homogeneous haze style from previous years, NTIRE2023 introduces 50 image pairs, each of a much higher resolution of 4000 $\times$ 6000. 
The increase in image size leads to larger volume of training data and greater demand in computation resources.
As the ground truth images of the 5 validation data and 5 test data are not public so far, we can only make use of the 40 train data when not evaluating on the server.
We adopt different train/test splitting strategies for performing quantitative comparisons of SOTA methods and ablation studies.
For methods comparison, we choose the first 35 images of the official training set as our training data, and the rest 5 are used for testing.
For ablation study, we use all 40 pairs to perform training, and get testing scores using the online validation server of the challenge.

\subsection{Implementation Details}
The input images are randomly cropped to a size of 256 $\times$ 256, and augmented by several data augmentation strategies, including 90, 180, 270 degrees of random rotation, horizontal flip, and vertical flip.
Note that, we do not apply any augmentation strategy related with brightness or color change as we have no intention to jeopardize the adjusted color distributions produced by our data-preprocessing method.
We use the AdamW \cite{loshchilov2017decoupled} ($\beta_1=0.9, \beta_2=0.999$) as our optimizer. 
The learning rate is initially set to $1e^{-4}$ and decreased to $1e^{-6}$ with a cosine annealing strategy.
We implement with the PyTorch library \cite{paszke2017automatic} on two Nvidia Titan XP GPUs.
Peak Signal to Noise Ratio (PSNR) and the Structural Similarity Index (SSIM) are used as two metrics for quantitative evaluation. 

\begin{table}[!t]
\small
\centering
\caption{Ablation study for architectures and data-preprocessing techniques. The scores are evaluated using NTIRE2023 online validation server.}
\setlength{\tabcolsep}{0.9mm}{}{
\begin{tabular}{*{10}{c|}}
\hline
\multicolumn{2}{|c|}{\multirow{2}*{\textbf{Data}}}& \multicolumn{2}{c|}{Res2Net+RCAN} & \multicolumn{2}{|c|}{Ours} \\
 \cline{3-6}
\multicolumn{2}{|c|}{} & \textbf{PSNR} & \textbf{SSIM} & \textbf{PSNR} & \textbf{SSIM}\\
\hline
\multicolumn{2}{|c|}{NH-HAZE23 only}& 20.68 & 0.678 &21.54 & 0.682  \\
\multicolumn{2}{|c|}{NH-HAZE20+21+23}&20.86  & 0.688  &21.54   &0.689 \\
\multicolumn{2}{|c|}{NH-HAZE20+21+23 GC}&21.08  & 0.690 &21.58  &0.693  \\
\multicolumn{2}{|c|}{NH-HAZE20+21+23 RGB} & \textbf{21.26}  & \textbf{0.693}  &\textbf{21.94}  & \textbf{0.697} \\
\hline
\end{tabular}}

\label{table:ab}
\end{table}

\begin{figure*}[!t]
\centering
\includegraphics[width=1.0\linewidth]{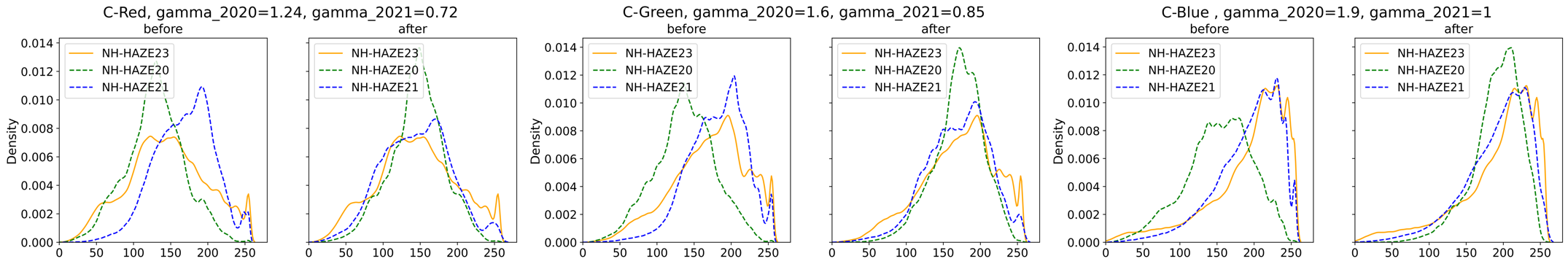}
\caption{ 
Comparison of RGB-wise distribution of datasets (hazy) before and after being processed by our proposed method.
}
\label{Fig.gamma_line_haze}
\end{figure*}

\subsection{Ablation Study}
We conduct comprehensive ablation studies to analyze and demonstrate the effectiveness of our data-preprocessing method and proposed network architecture.

\subsubsection{Importance of Data-Centric Design.}\label{sec:data_centric}
In Section \ref{sec:dpp}, we emphasize the importance of data for succeeding in non-homogeneous dehazing. To further demonstrate, we conduct experiments on several datasets with different data-preprocessing methods. 
There are four sets of training data, including: 1) \textbf{NH-HAZE23 only}: only the data from NTIRE2023 challenge is used; 2) \textbf{NH-HAZE20+21+23}: a direct combination of data from NH-HAZE20, NH-HAZE21 and NH-HAZE23 datasets;  3) \textbf{NH-HAZE20+21+23 GC}: a combination of data from NH-HAZE20, NH-HAZE21 and NH-HAZE23 with the GT data from NH-HAZE20 and 21 being processed by gray-scale gamma correction as \cite{yu2021two, Fu_2021_CVPR}; 4) \textbf{NH-HAZE20+21+23 RGB}:  a combination of data from NH-HAZE20, NH-HAZE21 and NH-HAZE23 with both the hazy and GT data from NH-HAZE20 and 21 being processed by our proposed approach described in Section \ref{sec:dpp}.
We employ these sets of data on two different model architectures.
The first model is from the NTIRE2021 challenge \cite{yu2021two} (we refer to as Res2Net+RCAN), where Res2Net \cite{gao2019res2net} is adopted as the backbone. 
The second model is our proposed one introduced in Section \ref{sec:model_arch}.
In total, we conduct 8 individual experiments, and report their best results (in terms of the PSNR index) evaluated on the NTIRE2023 online validation sever. The results are shown in Table \ref{table:ab}.

By comparing the first and second row of  Table \ref{table:ab}, we find that the direct combination of all the available data yields limited improvements for both our model and \cite{yu2021two}. 
By comparing the last two rows with the second row in Table \ref{table:ab}, it can be observed that performing data-preprocessing is generally beneficial. Not surprisingly, the models trained on our dataset achieve the best performance. These results  reinforce the importance of data-centric engineering.


To qualitative evaluate the importance of data engineering, we show in Figure \ref{Fig.ablation} by the images generated from the models respectively trained on \textbf{NH-HAZE20+21+23 GC} and \textbf{NH-HAZE20+21+23 RGB}. By comparing the two results with ground truth, it is obvious that the model trained on our processed dataset can generate more faithful results in terms of color and brightness. Specifically, the colors of the building and trees of ours are much more in line with those in the ground truth, while the compared one tend to generate green objects. 



\subsubsection{Effectiveness of Transformer} 
As noted in Section \ref{sec:model_arch}, our network is built upon the recent work \cite{yu2021two}. The main difference is that we replace the Res2Net branch in \cite{yu2021two} with the proposed Transformer-based structure. By quickly check the performance of our method and that of Res2Net+RCAN on four datasets in Table \ref{table:ab}, it can be easily observed that our method always outperforms Res2Net+RCAN by a significant margin. This illustrate the effectiveness of using Transformer in non-homogeneous dehazing.

\begin{figure}[!t]
\centering
\includegraphics[width=0.9\linewidth]{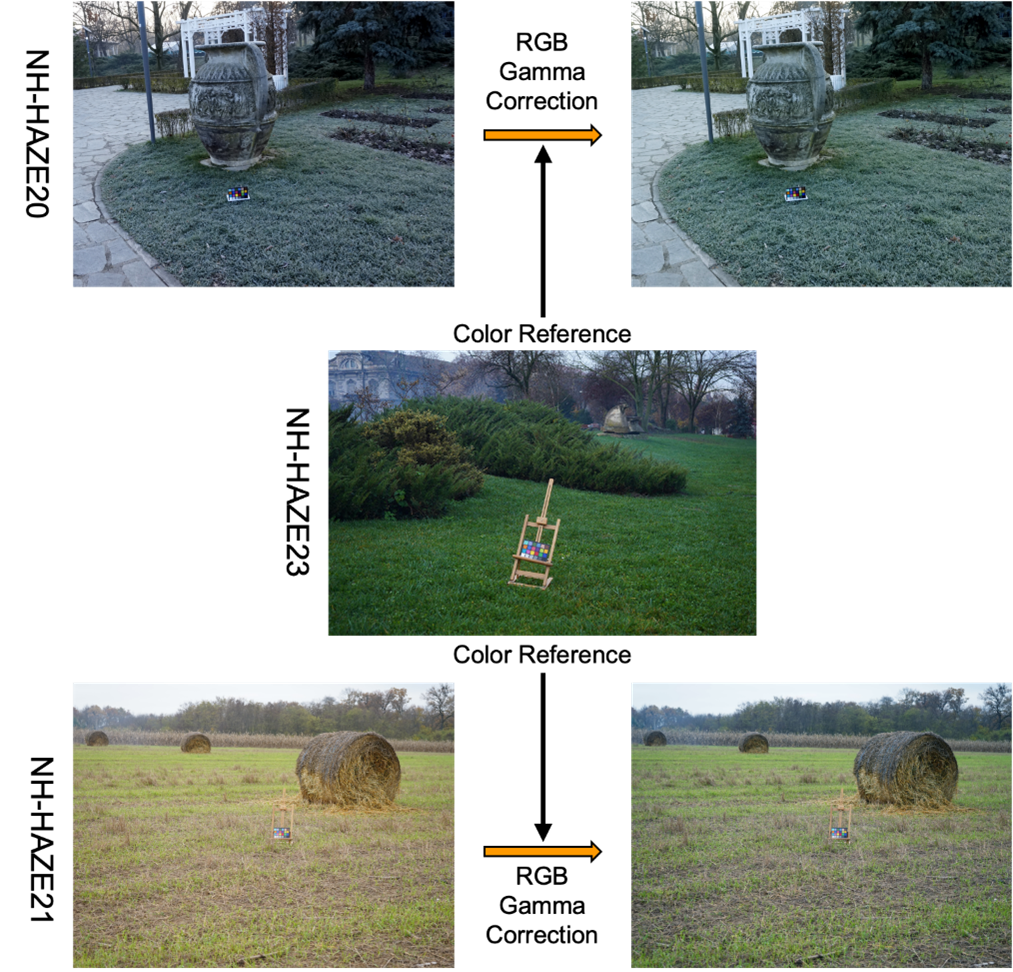}
\caption{ 
Examples from NH-HAZE20 and NH-HAZE21 datasets to visually showcase the color correction.
}
\label{Fig.gamma_example}
\end{figure}

\begin{figure*}[!t]
\centering
\includegraphics[width=1.0\textwidth]{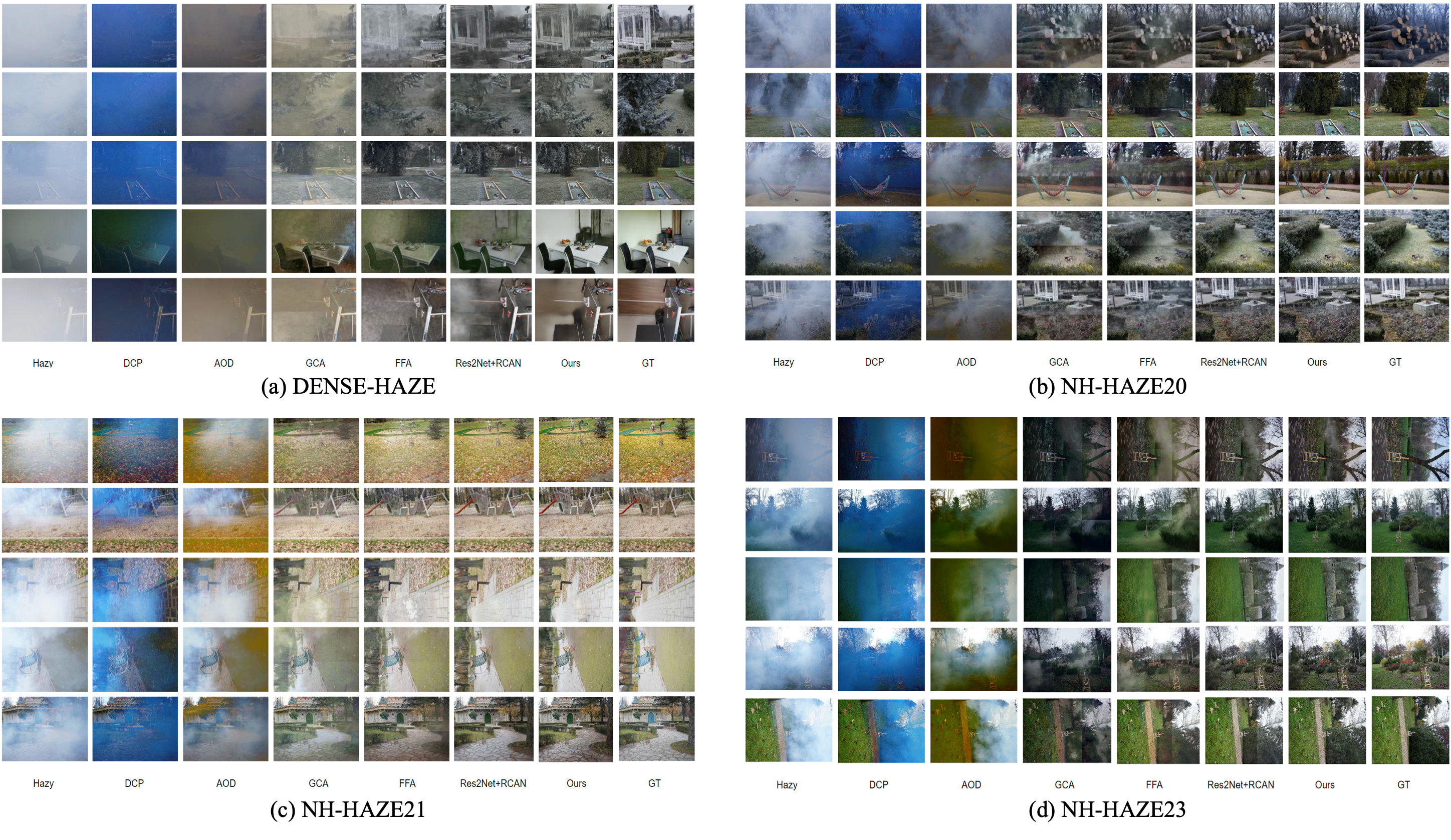}
\caption{Qualitative evaluation on the four representative datasets, i.e., DENSE-HAZE, NH-HAZE20, NH-HAZE21 and NH-HAZE23. For DENSE-HAZE and NH-HAZE20, we follow the official train, val and test split. For NH-HAZE21 and NH-HAZE23, due to the unavailable of test data, we split the released official training data to our training set and test set.
}
\label{Fig.comp_sota}
\end{figure*}

\begin{table*}[ht]
\centering
\caption{Quantitative evaluation on DENSE-HAZE, NH-HAZE20, NH-HAZE21 and NH-HAZE23 datasets. The best results are marked in \textbf{bold}, and the second bests are marked with \underline{underlines}.\label{table:comp_sota}}
\scalebox{1.0}{
\setlength{\tabcolsep}{2.6mm}{}{
\begin{tabular}{*{12}{c|}}
\hline
\multicolumn{2}{|c|}{\multirow{2}*{\textbf{Methods}}}& \multicolumn{2}{c|}{O-HAZE} & \multicolumn{2}{|c|}{DENSE-HAZE} & \multicolumn{2}{c|}{NH-HAZE20} & \multicolumn{2}{c|}{NH-HAZE21} & \multicolumn{2}{c|}{NH-HAZE23}\\ 
 \cline{3-12}
\multicolumn{2}{|c|}{} & \textbf{PSNR} & \textbf{SSIM} & \textbf{PSNR} & \textbf{SSIM} & \textbf{PSNR} & \textbf{SSIM} & \textbf{PSNR} & \textbf{SSIM} & \textbf{PSNR} & \multicolumn{1}{c|}{\textbf{SSIM}}\\
\hline
\multicolumn{2}{|c|}{DCP}& 12.92  & 0.505 & 10.85  & 0.404 & 12.29 & 0.411 & 11.30 & 0.605 & 11.87 & 0.470 \\
\multicolumn{2}{|c|}{AOD}& 17.69 & 0.616 & 13.30 & 0.469 & 13.44 & 0.413 & 13.22 & 0.613  & 12.47 & 0.369 \\
\multicolumn{2}{|c|}{GCANet}& 19.50 & 0.660 & 12.42 & 0.478 & 17.58 & 0.594 & 18.76 & 0.768  & 16.36 & 0.512 \\
\multicolumn{2}{|c|}{FFA} & 22.12 & 0.768 & 16.26 & 0.545 & 18.51 & 0.637 & 20.40 & 0.806 & 18.09 & 0.585 \\
\multicolumn{2}{|c|}{Res2Net+RCAN}& \underline{25.54} & \underline{0.783} & \textbf{16.36} & \textbf{0.582} & \textbf{21.44} & \underline{0.704} & \underline{21.66} & \textbf{0.843} & \underline{20.11}  & \underline{0.627}  \\
\multicolumn{2}{|c|}{Ours}& \textbf{25.98}  & \textbf{0.789}  & \underline{16.31}  & \underline{0.561}  & \textbf{21.44}   & \textbf{0.710}  & \textbf{21.67}  & \underline{0.838}  & \textbf{20.53}  & \textbf{0.636} \\
\hline
\end{tabular}}}

\end{table*}

\subsection{Further Analysis on Data-Centric Engineering}
In Figure \ref{Fig.clean_hist}, we show the distribution change of the ground truths after applying the proposed RGB gamma correction. 
In Figure \ref{Fig.gamma_line_haze}, we provide the distribution change of NH-HAZE20 and 21 hazy images as a supplement. 
It can be observed that after our data-preprocessing, the distribution of the three image channels (RGB) of the NH-HAZE20 and 21 hazy images are more in line with those of NH-HAZE23 data. Figure \ref{Fig.gamma_example} further qualitatively illustrates the images before and after data-preprocessing. The results shows that colors of the processed NH-HAZE20 and NH-HAZE21 data are much more similar to that of NH-HAZE23. We emphasize that this data-centric engineering is the key that helps our method stand out in the competition.
Based on both the analysis in this section and Section \ref{sec:data_centric}, we conclude that the data quality is one of the determining factors, possibly the most important one under the NTIRE dehazing challenges.

\subsection{Comparisons with the State-of-the-art Methods}
To conduct comparisons, we select five state-of-the-art methods, including DCP\cite{dcp}, AOD-Net \cite{li2017aod}, GCANet \cite{GCA}, FFA \cite{Qin_Wang_Bai_Xie_Jia_2020}, and Res2Net+RCAN \cite{yu2021two}.

In Table \ref{table:comp_sota}, we illustrate the best PSNR and SSIM indexes of each method on five different datasets.
The methods adopting the two-branch framework  perform generally well on all the datasets, where in Figure \ref{Fig.comp_sota}, Res2Net+RCAN and our method can produce visual pleasing results on all datasets.
They unveil significantly better performance when dealing with non-homogeneous haze patterns, as we could observe from the results on NH-HAZE20, 21 and 23.
Therefore, the two branch framework remains dominating in the limited data scenarios. 

It is worth noticing that our model substantially outperforms the Res2Net+RCAN model only on O-HAZE and NTIRE2023.
We argue the reason behind is that due to the huge increase of image resolution on O-HAZE and NH-HAZE23 datasets. For example, the number of pixels in NH-HAZE23 data is 6.25 times larger than that of the combination of NH-HAZE20 and NH-HAZE21 datasets.
Since our transformer-based model contains more learnable parameters, a larger training dataset can essentially alleviate the overfitting problem. 
This phenomenon further indicates that when it comes to a limited data setting, it is more critical to investigate in a data-centric manner other than simply improving the model's capacity.

%% file: conclusion.tex
\section{Conclusion}
In this paper, we propose a method targeting on non-homogeneous dehazing. 
It consists of a data-preprocessing strategy inspired by data-centric AI and a Transformer based two-branch model structure.
Combining them together, we construct a solution that outperforms the SOTA methods, which stimulates our advocation on treating the model and the data equally important. Additionally, extensive experimental results provide strong support to the effectiveness of our method.



%% file: main.bbl
\begin{thebibliography}{10}\itemsep=-1pt

\bibitem{dc_comp}
Data-centric ai competition submission guide, 2021.

\bibitem{ancuti2018ntire}
Cosmin Ancuti, Codruta~O Ancuti, and Radu Timofte.
\newblock Ntire 2018 challenge on image dehazing: Methods and results.
\newblock In {\em Proceedings of the IEEE Conference on Computer Vision and
  Pattern Recognition Workshops}, pages 891--901, 2018.

\bibitem{Dense-Haze_2019}
Codruta~O. Ancuti, Cosmin Ancuti, Mateu Sbert, and Radu Timofte.
\newblock Dense haze: A benchmark for image dehazing with dense-haze and
  haze-free images.
\newblock In {\em IEEE International Conference on Image Processing (ICIP)},
  IEEE ICIP 2019, 2019.

\bibitem{dense-haze_report}
C.~O. {Ancuti}, C. {Ancuti}, R. {Timofte}, L. {Van Gool}, L. {Zhang}, M.
  {Yang}, T. {Guo}, X. {Li}, V. {Cherukuri}, V. {Monga}, H. {Jiang}, S. {Yang},
  Y. {Liu}, X. {Qu}, P. {Wan}, D. {Park}, S.~Y. {Chun}, M. {Hong}, J. {Huang},
  Y. {Chen}, S. {Chen}, B. {Wang}, P.~N. {Michelini}, H. {Liu}, D. {Zhu}, J.
  {Liu}, S. {Santra}, R. {Mondal}, B. {Chanda}, P. {Morales}, T. {Klinghoffer},
  L.~M. {Quan}, Y. {Kim}, X. {Liang}, R. {Li}, J. {Pan}, J. {Tang}, K.
  {Purohit}, M. {Suin}, A.~N. {Rajagopalan}, R. {Schettini}, S. {Bianco}, F.
  {Piccoli}, C. {Cusano}, L. {Celona}, S. {Hwang}, Y.~S. {Ma}, H. {Byun}, S.
  {Murala}, A. {Dudhane}, H. {Aulakh}, T. {Zheng}, T. {Zhang}, W. {Qin}, R.
  {Zhou}, S. {Wang}, J. {Tarel}, C. {Wang}, and J. {Wu}.
\newblock Ntire 2019 image dehazing challenge report.
\newblock In {\em 2019 IEEE/CVF Conference on Computer Vision and Pattern
  Recognition Workshops (CVPRW)}, pages 2241--2253, 2019.

\bibitem{O-HAZE_2018}
Codruta~O. Ancuti, Cosmin Ancuti, Radu Timofte, and Christophe~De Vleeschouwer.
\newblock O-haze: a dehazing benchmark with real hazy and haze-free outdoor
  images.
\newblock In {\em IEEE Conference on Computer Vision and Pattern Recognition,
  NTIRE Workshop}, NTIRE CVPR'18, 2018.

\bibitem{ancuti2020ntire}
Codruta~O Ancuti, Cosmin Ancuti, Florin-Alexandru Vasluianu, and Radu Timofte.
\newblock Ntire 2020 challenge on nonhomogeneous dehazing.
\newblock In {\em Proceedings of the IEEE/CVF Conference on Computer Vision and
  Pattern Recognition Workshops}, pages 490--491, 2020.

\bibitem{ancuti2021ntire}
Codruta~O Ancuti, Cosmin Ancuti, Florin-Alexandru Vasluianu, and Radu Timofte.
\newblock Ntire 2021 nonhomogeneous dehazing challenge report.
\newblock In {\em Proceedings of the IEEE/CVF Conference on Computer Vision and
  Pattern Recognition}, pages 627--646, 2021.

\bibitem{ancuti2023ntire}
Codruta~O Ancuti, Cosmin Ancuti, Florin-Alexandru Vasluianu, and Radu Timofte.
\newblock Ntire 2023 challenge on nonhomogeneous dehazing.
\newblock In {\em Proceedings of the IEEE/CVF Conference on Computer Vision and
  Pattern Recognition Workshops}, 2023.

\bibitem{berman2016non}
Dana Berman, Shai Avidan, et~al.
\newblock Non-local image dehazing.
\newblock In {\em Proceedings of the IEEE conference on computer vision and
  pattern recognition}, pages 1674--1682, 2016.

\bibitem{cai2016dehazenet}
Bolun Cai, Xiangmin Xu, Kui Jia, Chunmei Qing, and Dacheng Tao.
\newblock Dehazenet: An end-to-end system for single image haze removal.
\newblock {\em IEEE Transactions on Image Processing}, 25(11):5187--5198, 2016.

\bibitem{GCA}
D. {Chen}, M. {He}, Q. {Fan}, J. {Liao}, L. {Zhang}, D. {Hou}, L. {Yuan}, and
  G. {Hua}.
\newblock Gated context aggregation network for image dehazing and deraining.
\newblock In {\em 2019 IEEE Winter Conference on Applications of Computer
  Vision (WACV)}, pages 1375--1383, 2019.

\bibitem{deng2009imagenet}
Jia Deng, Wei Dong, Richard Socher, Li-Jia Li, Kai Li, and Li Fei-Fei.
\newblock Imagenet: A large-scale hierarchical image database.
\newblock In {\em 2009 IEEE conference on computer vision and pattern
  recognition}, pages 248--255. Ieee, 2009.

\bibitem{dosovitskiy2020image}
Alexey Dosovitskiy, Lucas Beyer, Alexander Kolesnikov, Dirk Weissenborn,
  Xiaohua Zhai, Thomas Unterthiner, Mostafa Dehghani, Matthias Minderer, Georg
  Heigold, Sylvain Gelly, et~al.
\newblock An image is worth 16x16 words: Transformers for image recognition at
  scale.
\newblock {\em arXiv preprint arXiv:2010.11929}, 2020.

\bibitem{fattal2014dehazing}
Raanan Fattal.
\newblock Dehazing using color-lines.
\newblock {\em ACM transactions on graphics (TOG)}, 34(1):1--14, 2014.

\bibitem{Fu_2021_CVPR}
Minghan Fu, Huan Liu, Yankun Yu, Jun Chen, and Keyan Wang.
\newblock Dw-gan: A discrete wavelet transform gan for nonhomogeneous dehazing.
\newblock In {\em Proceedings of the IEEE/CVF Conference on Computer Vision and
  Pattern Recognition (CVPR) Workshops}, pages 203--212, June 2021.

\bibitem{gao2019res2net}
Shanghua Gao, Ming-Ming Cheng, Kai Zhao, Xin-Yu Zhang, Ming-Hsuan Yang, and
  Philip~HS Torr.
\newblock Res2net: A new multi-scale backbone architecture.
\newblock {\em IEEE transactions on pattern analysis and machine intelligence},
  2019.

\bibitem{Girshick_2015_ICCV}
Ross Girshick.
\newblock Fast r-cnn.
\newblock In {\em Proceedings of the IEEE International Conference on Computer
  Vision (ICCV)}, December 2015.

\bibitem{dcp}
Kaiming He, Jian Sun, and Xiaoou Tang.
\newblock Single image haze removal using dark channel prior.
\newblock {\em IEEE transactions on pattern analysis and machine intelligence},
  33(12):2341--2353, 2010.

\bibitem{jacobsadaptive}
Robert~A Jacobs, Michael~I Jordan, Steven~J Nowlan, and Geoffrey~E Hinton.
\newblock Adaptive mixtures of local experts.
\newblock {\em Neural computation}, 3(1):79--87, 1991.

\bibitem{Kornblith_2019_CVPR}
Simon Kornblith, Jonathon Shlens, and Quoc~V. Le.
\newblock Do better imagenet models transfer better?
\newblock In {\em Proceedings of the IEEE/CVF Conference on Computer Vision and
  Pattern Recognition (CVPR)}, June 2019.

\bibitem{li2017aod}
Boyi Li, Xiulian Peng, Zhangyang Wang, Jizheng Xu, and Dan Feng.
\newblock Aod-net: All-in-one dehazing network.
\newblock In {\em Proceedings of the IEEE international conference on computer
  vision}, pages 4770--4778, 2017.

\bibitem{liu2022towards}
Huan Liu, Zijun Wu, Liangyan Li, Sadaf Salehkalaibar, Jun Chen, and Keyan Wang.
\newblock Towards multi-domain single image dehazing via test-time training.
\newblock In {\em Proceedings of the IEEE/CVF Conference on Computer Vision and
  Pattern Recognition}, pages 5831--5840, 2022.

\bibitem{Liu_2020_CVPR_Workshops}
Jing Liu, Haiyan Wu, Yuan Xie, Yanyun Qu, and Lizhuang Ma.
\newblock Trident dehazing network.
\newblock In {\em Proceedings of the IEEE/CVF Conference on Computer Vision and
  Pattern Recognition (CVPR) Workshops}, June 2020.

\bibitem{liu2022swin}
Ze Liu, Han Hu, Yutong Lin, Zhuliang Yao, Zhenda Xie, Yixuan Wei, Jia Ning, Yue
  Cao, Zheng Zhang, Li Dong, et~al.
\newblock Swin transformer v2: Scaling up capacity and resolution.
\newblock In {\em Proceedings of the IEEE/CVF conference on computer vision and
  pattern recognition}, pages 12009--12019, 2022.

\bibitem{loshchilov2017decoupled}
Ilya Loshchilov and Frank Hutter.
\newblock Decoupled weight decay regularization.
\newblock {\em arXiv preprint arXiv:1711.05101}, 2017.

\bibitem{middleton1952vision}
William Edgar~Knowles Middleton.
\newblock {\em Vision through the atmosphere}.
\newblock University of Toronto Press, 1952.

\bibitem{motamedi2021datacentric}
Mohammad Motamedi, Nikolay Sakharnykh, and Tim Kaldewey.
\newblock A data-centric approach for training deep neural networks with less
  data, 2021.

\bibitem{paszke2017automatic}
Adam Paszke, Sam Gross, Soumith Chintala, Gregory Chanan, Edward Yang, Zachary
  DeVito, Zeming Lin, Alban Desmaison, Luca Antiga, and Adam Lerer.
\newblock Automatic differentiation in pytorch.
\newblock 2017.

\bibitem{Qin_Wang_Bai_Xie_Jia_2020}
Xu Qin, Zhilin Wang, Yuanchao Bai, Xiaodong Xie, and Huizhu Jia.
\newblock Ffa-net: Feature fusion attention network for single image dehazing.
\newblock {\em Proceedings of the AAAI Conference on Artificial Intelligence},
  34(07):11908--11915, Apr. 2020.

\bibitem{ren2018gated}
Wenqi Ren, Lin Ma, Jiawei Zhang, Jinshan Pan, Xiaochun Cao, Wei Liu, and
  Ming-Hsuan Yang.
\newblock Gated fusion network for single image dehazing.
\newblock In {\em Proceedings of the IEEE Conference on Computer Vision and
  Pattern Recognition}, pages 3253--3261, 2018.

\bibitem{shao2020domain}
Yuanjie Shao, Lerenhan Li, Wenqi Ren, Changxin Gao, and Nong Sang.
\newblock Domain adaptation for image dehazing.
\newblock In {\em Proceedings of the IEEE/CVF Conference on Computer Vision and
  Pattern Recognition}, pages 2808--2817, 2020.

\bibitem{simonyan2014very}
Karen Simonyan and Andrew Zisserman.
\newblock Very deep convolutional networks for large-scale image recognition.
\newblock {\em arXiv preprint arXiv:1409.1556}, 2014.

\bibitem{tan2018survey}
Chuanqi Tan, Fuchun Sun, Tao Kong, Wenchang Zhang, Chao Yang, and Chunfang Liu.
\newblock A survey on deep transfer learning, 2018.

\bibitem{tan2008visibility}
Robby~T Tan.
\newblock Visibility in bad weather from a single image.
\newblock In {\em 2008 IEEE conference on computer vision and pattern
  recognition}, pages 1--8. IEEE, 2008.

\bibitem{Wu_2020_CVPR_Workshops}
Haiyan Wu, Jing Liu, Yuan Xie, Yanyun Qu, and Lizhuang Ma.
\newblock Knowledge transfer dehazing network for nonhomogeneous dehazing.
\newblock In {\em Proceedings of the IEEE/CVF Conference on Computer Vision and
  Pattern Recognition (CVPR) Workshops}, June 2020.

\bibitem{wu2020knowledge}
Haiyan Wu, Jing Liu, Yuan Xie, Yanyun Qu, and Lizhuang Ma.
\newblock Knowledge transfer dehazing network for nonhomogeneous dehazing.
\newblock In {\em Proceedings of the IEEE/CVF conference on computer vision and
  pattern recognition workshops}, pages 478--479, 2020.

\bibitem{yu2021two}
Yankun Yu, Huan Liu, Minghan Fu, Jun Chen, Xiyao Wang, and Keyan Wang.
\newblock A two-branch neural network for non-homogeneous dehazing via ensemble
  learning.
\newblock In {\em Proceedings of the IEEE/CVF conference on computer vision and
  pattern recognition}, pages 193--202, 2021.

\bibitem{zha2023datacentric}
Daochen Zha, Zaid~Pervaiz Bhat, Kwei-Herng Lai, Fan Yang, and Xia Hu.
\newblock Data-centric ai: Perspectives and challenges, 2023.

\bibitem{zhang2018densely}
He Zhang and Vishal~M Patel.
\newblock Densely connected pyramid dehazing network.
\newblock In {\em Proceedings of the IEEE conference on computer vision and
  pattern recognition}, pages 3194--3203, 2018.

\bibitem{zhang2018image}
Yulun Zhang, Kunpeng Li, Kai Li, Lichen Wang, Bineng Zhong, and Yun Fu.
\newblock Image super-resolution using very deep residual channel attention
  networks.
\newblock In {\em Proceedings of the European conference on computer vision
  (ECCV)}, pages 286--301, 2018.

\bibitem{zhao2016loss}
Hang Zhao, Orazio Gallo, Iuri Frosio, and Jan Kautz.
\newblock Loss functions for image restoration with neural networks.
\newblock {\em IEEE Transactions on computational imaging}, 3(1):47--57, 2016.

\bibitem{zhou2020cggan}
Zhaorun Zhou, Zhenghao Shi, Mingtao Guo, Yaning Feng, and Minghua Zhao.
\newblock Cggan: A context guided generative adversarial network for single
  image dehazing, 2020.

\bibitem{zhu2017unpaired}
Jun-Yan Zhu, Taesung Park, Phillip Isola, and Alexei~A Efros.
\newblock Unpaired image-to-image translation using cycle-consistent
  adversarial networks.
\newblock In {\em Proceedings of the IEEE international conference on computer
  vision}, pages 2223--2232, 2017.

\bibitem{zhu2015fast}
Qingsong Zhu, Jiaming Mai, and Ling Shao.
\newblock A fast single image haze removal algorithm using color attenuation
  prior.
\newblock {\em IEEE transactions on image processing}, 24(11):3522--3533, 2015.

\end{thebibliography}
